\pdfoutput=1
\documentclass[11pt,a4paper]{article}
\usepackage[hyperref]{acl2021}
\usepackage{times}
\usepackage{graphicx}
\usepackage{latexsym}

\usepackage{microtype}

\aclfinalcopy 
\def\aclpaperid{2335} 

\usepackage{xcolor}
\usepackage{multirow} 
\usepackage{caption}
\usepackage{subcaption}
\usepackage[stable]{footmisc}
\usepackage{makecell}

\definecolor{nadavgreen}{rgb}{0.158, 0.688, 0.178}

\usepackage{arydshln}
\usepackage{color}
\usepackage{xspace}
\usepackage{paralist}

\newcommand{\remove}[1]{}
\newcommand{\MLP}{Iggy\xspace}
\newcommand{\xhdr}[1]{\vspace{1mm}\noindent{{\bf #1.}}}

\newcommand\rurl[1]{%
  \href{http://#1}{\nolinkurl{#1}}%
}
\title{How Did This Get Funded?!\\
Automatically Identifying Quirky Scientific Achievements}

\author{Chen Shani\thanks{ \ \space Equal contribution}  , \space Nadav Borenstein\footnotemark[1]  , \space Dafna Shahaf \\ \quad 
 The Hebrew University of Jerusalem \\
  \texttt{\{chenxshani, nadav.borenstein, dshahaf\}@cs.huji.ac.il} \\
}

 \date{}

\begin{document}
\maketitle

\begin{abstract}
\remove{
Humor is an essential part of being human, and has been studied for millennia. Studying humor through the lens of computer science (and in particular, automated analysis of humor) has important implications for artificial intelligence and human-computer interaction.
In this paper we introduce a novel problem in humor mining -- automatically detecting funny and unusual scientific papers. We are inspired by the Ig Nobel Prize, a satirical prize awarded annually to celebrate funny scientific achievements (example past winner: ``Are Cows More Likely to Lie Down the Longer They Stand?"). We construct a dataset containing thousands of funny scientific papers and use it to learn classifiers, combining findings from psychology and linguistics with recent advances in NLP.
We use our classifiers to identify potentially funny papers in a large dataset of $630$,$000$ papers. Our models achieved NDCG scores of $0.68$-$0.74$, demonstrating the utility of our approach.

}

Humor is an important social phenomenon, serving complex social and psychological functions.
However, despite being studied for millennia humor is computationally not well understood, often considered an AI-complete problem.

In this work, we introduce a novel setting in humor mining: automatically detecting funny and unusual scientific papers. We are inspired by the Ig Nobel prize, a satirical prize awarded annually to celebrate funny scientific achievements (example past winner: ``Are cows more likely to lie down the longer they stand?"). This challenging task has unique characteristics that make it particularly suitable for automatic learning.


We construct a dataset containing thousands of funny papers and use it to learn classifiers, combining findings from psychology and linguistics with recent advances in NLP. We use our models to identify potentially funny papers in a large dataset of over $630$,$000$ articles. The results demonstrate the potential of our methods, and more broadly the utility of integrating state-of-the-art NLP methods with insights from more traditional disciplines. 
\end{abstract}

\section{Introduction}
\label{sec:intro}
Humor is an important aspect of the way we interact with each other, serving complex social functions \citep{martineau1972model}. Humor can function either as a lubricant or as an abrasive: it can be used as a key for improving interpersonal relations and building trust \citep{wanzer1996funny, wen2015omg}, or help us work through difficult topics. It can also aid in breaking taboos and holding power to account. Enhancing the humor capabilities of computers has tremendous potential to better understand interactions between people, as well as build more natural human-computer interfaces. 

Nevertheless, computational humor remains a long-standing challenge in AI; It requires complex language understanding, manipulation capabilities, creativity, common sense, and empathy. Some even claim that computational humor is an AI-complete problem \cite{stock2002hahacronym}.

As humor is a broad phenomenon, most works on computational humor focus on \emph{specific} humor types, such as knock-knock jokes or one-liners \cite{mihalcea2006learning,taylor2004computationally}.
In this work, we present a novel humor recognition task: identifying quirky, funny scientific contributions. We are inspired by the {\bf Ig Nobel prize}\footnote{\rurl{improbable.com/ig-about}}, a satiric prize awarded annually to ten scientific achievements that ``first make people laugh, and then think''. Past Ig Nobel winners include ``Chickens prefer beautiful humans'' and ``Beauty is in the eye of the beer holder: People who think they are drunk also think they are attractive''. 

Automatically identifying candidates for the Ig Nobel prize provides a unique perspective on humor. 
Unlike most humor recognition tasks, the humor involved is sophisticated, and requires common sense, as well as specialized knowledge and understanding of the scientific culture. 
On the other hand, this task has several characteristics rendering it attractive: the funniness of the paper can often be recognized from its \emph{title} alone, which is short, with simple syntax and no complex narrative structure (as opposed to longer jokes). Thus, this is a relatively clean setting to explore our methods.

We believe humor in science is also particularly interesting to explore, as humor is strongly tied to creativity. Quirky contributions could sometimes indicate fresh perspectives and pioneering attempts to expand the frontiers of science. For example, Andre Geim won an Ig Nobel in $2000$ for levitating a frog using magnets and a Nobel Prize in Physics in $2010$. The Nobel committee explicitly attributed the win to his playfulness \citep{NobelPressRelease}.





Our contributions are:



\begin{compactitem}
\item We formulate a novel humor recognition task in the scientific domain. 
\item We construct a dataset containing thousands of funny scientific papers. 
\item We develop multiple classifiers, combining findings from psychology and linguistics with recent NLP advances. We evaluate them both on our dataset and in a real-world setting, identifying potential Ig Nobel candidates in a large corpus of over $0.6$M papers. 
\item We devise a rigorous, data-driven way to aggregate crowd workers' annotations for \textit{subjective} questions.
\item We release data and code\footnote{\rurl{github.com/nadavborenstein/Iggy}\label{github}}.
\end{compactitem}

Beyond the tongue-in-cheek nature of our application, we more broadly wish to promote combining data-driven research with more-traditional works in areas such as psychology. We believe insights from such fields could complement machine learning models, improving performance as well as enriching our understanding of the problem.

\section{Related Work}
\label{sec:related_work}

\xhdr{Humor in the Humanities}
A large body of theoretical work on humor stems from linguistics and psychology. \citet{ruch1992assessment} divided humor into three categories: incongruity, sexual, and nonsense (and created a three-dimensional humor test to account for them). Since our task is to detect humor in scientific contributions, we believe that the third category can be neglected under the assumption that no-nonsense article would (or at least, should) be published (notable exception: the Sokal hoax \cite{sokal1996transgressing}). 

The first category, incongruity, was first fully conceptualized by Kant in the eighteenth century \cite{shaw2010philosophy}. The well-agreed extensions to incongruity theory are the linguistics incongruity resolution model and semantic script theory of humor \cite{suls1972two, raskin1985semantic}. Both state that if a situation ended in a manner that contradicted our prediction (in our case, the title contains an unexpected term) and there exists a different, less likely rule to explain it -- the result is a humorous experience. Simply put, the source of humor lies in violation of expectations. 
Example Ig Nobel winners include: ``Will humans swim faster or slower in syrup?"  and "Coordination modes in the multisegmental dynamics of hula hooping".
   
The second category, sex-related humor is also common among Ig Nobel winning papers. Examples include: ``Effect of different types of textiles on sexual activity. Experimental study'' and ``Magnetic resonance imaging of male and female genitals during coitus and female sexual arousal".


\xhdr{Humor Detection in AI}
Most computational humor detection work done in the context of AI relies on supervised or semi-supervised methods and focuses on specific, narrow, types of jokes or humor. 

Humor detection is usually formulated as a binary text classification problem. Example domains include knock-knock jokes \citep{taylor2004computationally}, one-liners \citep{miller2017semeval, simpson2019predicting, liu2018modeling, mihalcea2005making, blinov2019large, mihalcea2006learning}, humorous tweets \citep{maronikolakis2020analyzing, donahue2017humorhawk, ortega2018uo, zhang2014recognizing}, humorous product reviews \citep{ziser2020humor, reyes2012making}, TV sitcoms \cite{bertero2016long}, short stories \citep{wilmot-keller-2020-modelling}, cartoons captions \citep{shahaf2015inside}, and even ``That's what she said" jokes \citep{hossain2017filling, kiddon2011s}. Related tasks such as irony, sarcasm and satire have also been explored in similarly narrow domains
\citep{davidov2010semi, reyes2012humor, ptavcek2014sarcasm}. 

\section{Problem Formulation and Dataset}
\label{sec:problem}
\label{sec:dataset}
Our goal in this paper is to automatically identify candidates for the Ig Nobel prize. More precisely, to automatically detect humor in scientific papers. 

First, we consider the question of input to our algorithm. 
\citet{sagi2008amusing} found a strong correlation between funny \emph{title} and humorous \emph{subject} in scientific papers. Motivated by this correlation, we manually inspected a subset of Ig Nobel winners. For the vast majority of them, reading the title was enough to determine whether it is funny; very rarely did we need to read the abstract, let alone the full paper. Typical past winners' titles include ``Why do old men have big ears?'' and ``If you drop it, should you eat it? Scientists weigh in on the 5-second rule". An example of a non-informative title is ``Pouring flows'', a paper calculating the optimal way to dunk a biscuit in a cup of tea. 

Based on this observation, we decided to focus on the papers' \emph{titles}. More formally: Given a title $t$ of an article, our goal is to learn a binary function $\varphi(t)\rightarrow \{0, 1\}$, reflecting whether the paper is humorous, or `Ig Nobel-worthy'. The main challenge, of course, lies in the construction of $\varphi$. 

To take a data-driven approach to tackle this problem, we crafted a first-of-its-kind dataset containing titles of funny scientific papers\footref{github}. We started from the $211$ Ig Nobel winners. Next, we manually collected humorous papers from online forums and blogs
\footnote{\raggedright E.g., \rurl{reddit.com/r/ScienceHumour}, \rurl{popsci.com/read/funny-science-blog}, \rurl{goodsciencewriting.wordpress.com}}, resulting in $1$,$707$ papers. 
We manually verified all of these papers can be used as positive examples. In Section \ref{sec:preResults} we give more indication these papers are indeed useful for our task.

For negative examples, we randomly sampled $1$,$707$ titles from Semantic Scholar\footnote{\rurl{api.semanticscholar.org/corpus/}} (to obtain a balanced dataset). 
We then classify each paper into one of the following scientific fields: neuroscience, medicine, biology, or exact sciences\footnote{Using \rurl{scimagojr.com} to map venues to fields.}. We balanced the dataset in a per-field manner. 
While some of these randomly sampled papers could, in principle, be funny, the vast majority of scientific papers are not (we validated this assumption through sampling).

\section{Humor-Theory Inspired Features}
\label{sec:analysis}
In deep learning, architecture engineering largely took the place of feature engineering.
One of the goals of our work is to evaluate the value of features inspired by domain experts.
In this section, we describe and formalize $127$ features implementing insights from humor literature. 
To validate the predictive power of the features that require training, we divide our data to train and test sets ($80\%/20\%$). We now describe the four major feature families.

\subsection{Unexpected Language}

Research suggests that \emph{surprise} is an important source of humor \cite{raskin1985semantic, suls1972two}. Indeed, we notice that titles of Ig Nobel winners often include an unexpected term or unusual language, e.g.: ``On the rheology of \textit{cats}'', ``Effect of \textit{coke} on sperm motility''
and ``\textit{Pigeons'} discrimination of \textit{paintings} by Monet and Picasso''.
To quantify unexpectedness, we create several different language-models (LMs):

\noindent \textbf{N-gram Based LMs.\ }
We train simple N-gram LMs with $n \in \{1, 2, 3\}$ on two corpora -- $630$,$000$ titles from Semantic Scholar, and $231$,$600$ one-line jokes \cite{shortjokes}. 
    
\noindent \textbf{Syntax-Based LMs.\ }
Here we test the hypothesis that humorous text has more surprising \emph{grammatical structure} \citep{oaks1994creating}. 
We replace each word in our Semantic Scholar corpus with its corresponding part-of-speech (POS) tag\footnote{Obtained using NLTK (\rurl{nltk.org})}. We then trained N-gram based LMs ($n \in \{1, 2, 3\}$) on this corpus.
    
\noindent \textbf{Transformer-Based LMs.\ }
We use three different Transformers based \cite{vaswani2017attention} models: 1) BERT \cite{devlin2018bert} (pre-trained on Wikipedia and the BookCorpus), 2) SciBERT \cite{beltagy2019scibert}, a variant of BERT optimized on scientific text from Semantic Scholar, and 3) GPT-2 \cite{radford2019language}, a large Transformer-based LM, trained on a dataset of $8$M web pages. We fine-tuned GPT-2 on our Semantic Scholar corpora (details in Appendix \ref{app:GPT2}). 


\noindent \textbf{Using the LMs.\ }
For each word in a title, we compute the word's perplexity. 
For the N-gram LMs and GPT-2, we compute the probability to see the word given the previous words in the sentence ($n-1$ previous words in the case of the N-gram models and all the previous words in the case of GPT-2). For the BERT-based models, we compute the masked loss of the word given the sentence. For each title, we computed the mean, maximum, and variance of the perplexity across all words in the title.

\subsection{Simple Language}
Inspired by previous findings \cite{ruch1992assessment, gultchin2019humor}, we hypothesize that titles of funny papers tend to be simpler (e.g., the past Ig Nobel winners: ``Chickens prefer beautiful humans" and ``Walking with coffee: Why does it spill?"). 
We utilize several simplicity measures: 

\noindent \textbf{Length.\ }
Short titles and titles containing many short words tend to be simpler. We compute title length and word lengths (mean, maximum, and variance of word lengths in the title).
    
\noindent \textbf{Readability.\ }
We use the automated readability index \cite{smith1967automated}.
    
\noindent \textbf{Age of Acquisition (AoA).\ }
A well-established measure for word's difficulty in psychology \cite{brysbaert2017test}, denoting word's difficulty by the age a child acquires it. We compute mean, maximum and variance AoA.
    
\noindent \textbf{AoA and Perplexity.\ }
Many basic words can be found in serious titles (e.g., `water' in a hydraulics paper). Funny titles, however, contain simple words which are also \emph{unexpected}. Thus, we combine AoA with perplexity. We compute word perplexity using the Semantic Scholar N-gram LMs and divide it by AoA. Higher values correspond to simpler and unexpected words. We compute the mean, maximum, minimum, and variance. 

\subsection{Crude Language}
According to relief theory, crude and scatological connotations are often considered humorous \cite{shurcliff1968judged} (e.g., the Ig Nobel winners ``Duration of urination does not change with body size'', ``Acute management of the zipper-entrapped penis'').

We trained a Naive Bayes SVM \cite{wang2012baselines} classifier over a dataset of toxic and rude Wikipedia comments \cite{toxic}, and compute title probability to be crude.
Similar to the AoA feature, we believe that crude words should also be unexpected to be considered funny. 
As before, we divide perplexity by the word's probability of being benign. Higher values correspond to crude and unexpected words. We compute the mean, maximum, minimum, and variance. 

\subsection{Funny Language}
\label{sec:funnyL}
Some words (e.g., nincompoop, razzmatazz) are inherently funnier than others (due to various reasons surveyed by \citet{gultchin2019humor}). It is reasonable that the funniness of a title is correlated with the funniness of its words. 
We measure funniness using the model of \citet{westbury2019wriggly}, quantifying noun funniness based on humor theories and human ratings. We measure the funniness of each noun in a title. We also multiplied perplexity and funniness (for funny and unexpected) and use the mean, maximum, minimum, and variance. 


\subsection{Feature Importance}

As a first reality check, we plotted the distribution of our features between funny and not-funny papers (see Appendix \ref{app:dataAnalysis} for representative examples). For example, we hypothesized that titles of funny papers might be linguistically similar to one-liners, and indeed we saw that the one-liner LM assigns lower perplexity to funny papers. Similarly, we saw a difference between the readability scores.

To measure the predictive power of our literature-inspired features, we use the Wilcoxon signed-rank test\footnote{A non-parametric paired difference test used to assess whether the mean ranks of two related samples differ.} (see Table \ref{tab:wilcoxon}). 
Interestingly, all feature families include useful features. Combining perplexity with other features (e.g., surprising and simple words) was especially prominent. In the next sections, we describe how we use those features to train models for detecting Ig Nobel worthy papers.


\section{Models}
\label{sec:model}
\begin{table}[t!]
 \begin{tabular}{c|c|c}
  \multirow{2}{*}{\makecell{Feature}} & \multirow{2}{*}{\makecell{Wilcoxon\\value}} & \multirow{2}{*}{P-value} \\ 
  &&\\
  \hline \hline
  \multicolumn{3}{c}{\textbf{Unexpected Language}} \\ \hline
  \makecell{Avg. Semantic Scholar\\2-gram LM} & 4850 & 3.6e-39 \\ \hline
  \makecell{Avg. POS 2-gram LM} & 18926 & 3e-7 \\ \hline
  \makecell{Avg. one-liners\\2-gram LM} & 6919 & 9.2e-33 \\ \hline
  \makecell{Avg. GPT-2 LM} & 7421 & 2.7e-31 \\ \hline
  \makecell{Avg. BERT LM} & 17153 & 9e-10 \\ \hline
  \multicolumn{3}{c}{\textbf{Simple Language}} \\ \hline
  \makecell{Readability} & 8931 & 8.6e-26 \\ \hline
  \makecell{Title's length} & 18493 & 2.4e-6 \\ \hline
  \makecell{Avg. AoA values} & 16768 & 2.2e-10 \\ \hline
  \makecell{Avg. AoA +\\2-gram LM} & 4882 & 4.6e-39 \\ \hline
  \multicolumn{3}{c}{\textbf{Crude Language}} \\ \hline
  \makecell{Crudeness classifier} & 17423 & 2.3e-9 \\ \hline
  \makecell{Avg. crudeness +\\2-gram LM} & 4755 & 1.8e-39 \\ \hline
  \multicolumn{3}{c}{\textbf{Funny Language}} \\ \hline
  \makecell{Avg. funny\\nouns model} & 20101 & 1.7e-5 \\ \hline
  \makecell{Avg. funny nouns +\\2-gram LM} & 8886 & 3.2e-27 \\ 
\end{tabular}
 \caption{Wilcoxon and p-values for representative features using our dataset (tested differentiating ability between funny and serious papers). Combining perplexity with other features seems particularly beneficial.}
  \label{tab:wilcoxon}
\end{table}

We can now create models to automatically detect scientific humor. As mentioned in Section \ref{sec:analysis}, one of our goals in this paper is to compare between the NLP SOTA huge-models approach and the literature-inspired approach. 
Thus, we trained a binary multi-layer perceptron (MLP) classifier using our dataset (described in Section \ref{sec:dataset}, see reproducibility details in Appendix \ref{app:MLP}), receiving as input the $127$ features from Section \ref{sec:analysis}. We named this classifier {\bf `{\MLP}'}, after the Ig Nobel prize.

As baselines representing the contemporary NLP approach (requiring huge compute and training data), we used {\bf BERT} \citep{devlin2018bert} and {\bf SciBERT} \citep{beltagy2019scibert}, which is a BERT variant optimized on scientific corpora, rendering it potentially more relevant for our task.
We fine-tuned SciBERT and BERT for Ig Nobel classification using our dataset (see Appendix \ref{app:finetune_bert} for implementation details). 

We also experimented with two models combining {\bf BERT\slash SciBERT with our features} (see Figure \ref{fig:modelflow} in Appendix \ref{app:combined_model}), denoted as {BERT}$^f$\slash {SciBERT}$^f$. In the spirit of the original BERT paper, we added two linear layers on top of the models and used a standard cross-entropy loss. The input to this final MLP is the concatenation of two vectors: our features' embedding and the last hidden vector from BERT\slash SciBERT ([CLS]). See Appendix \ref{app:combined_model} for implementation details.


For the sake of completeness, we note that we also conducted exploratory experiments with simple syntactic baselines (title length, maximal word length, title containing a question, title containing a colon) as well as BERT trained on sarcasm detection\footnote{\rurl{kaggle.com/raghavkhemka/sarcasm-detection-using-bert-92-accuracy}}. None of these baselines was strong enough on its own. We note that the colon-baseline tended to catch smart-aleck titles, but the topic was not necessarily funny. The sarcasm baseline achieved near guess-level accuracy ($0.482$), emphasizing the distinction between the two humor tasks.

\section{Evaluation on the Dataset}
\label{sec:preResults}
We first evaluate the five models (\MLP, SciBERT, BERT, {SciBERT}$^f$ and {BERT}$^f$) on our labeled dataset in terms of general accuracy and Ig Nobel retrieval ability. As naive baselines, we added two bag of words (BoW) based classifies: random forest (RF) and logistic regression (LR).

\xhdr{Accuracy}
We randomly split the dataset to train, development, and test sets ($80\%-10\%-10\%$), and used the development set to tune hyper-parameters (e.g., learning rate, number of training epochs). Table \ref{tab:our_dataset} summarizes the results. We note that all five models achieve very high accuracy scores and that the simple BoW models fall behind. This gives some indication about the inherent difficulty of the task. Both features-based \MLP and BERT-based models outperform simple baseline.
{SciBERT}$^f$ outperforms the other models across all measures.

\xhdr{Ig Nobel Winners Retrieval}
Our positive examples consist of $211$ Ig Nobel winners and additional $1$,$496$ humorous papers found on the web. Thus, the portion of real Ig Nobel winning papers in our data is relatively small. 
We now measure whether our web-originated papers serve as a good proxy for Ig Nobel winners. 
Thus, we split the dataset differently: the test set consists of the $211$ Ig Nobel winners, plus a random sample of $211$ negative titles (slightly increasing the test set size to $12$\%). Train set consists of the remaining $2$,$992$ papers. 
This experiment follows our initial inspiration of finding Ig Nobel-worthy papers, as we test our models' ability to retrieve only the real winners.

Table \ref{tab:ig_ret} demonstrate that our web-based funny papers are indeed a good proxy for Ig Nobel winners. 
Similar to the previous experiment, the combination of SOTA pretrained models with literature based features is superior. 

Based on both experiments, we conclude that our features are indeed informative for our Ig Nobel-worthy papers detection task.

\begin{table}
 \begin{tabular}{l|c|c|c}
  Model & Accuracy & Precision & Recall \\ 
  \hline 
  {\MLP}  & 0.897 & 0.901 & 0.893 \\ \hdashline %
  SciBERT & 0.910 & 0.911 & 0.911 \\ \hdashline %
  \makecell[l]{{SciBERT}$^f$} & \textbf{0.922} & \textbf{0.919} & \textbf{0.926} \\ \hdashline %
  BERT & 0.904 & 0.906 & 0.893 \\ \hdashline %
  \makecell[l]{{BERT}$^f$} & 0.900 & 0.899 & 0.902 \\ \hdashline %
  RF & 0.761 & 0.746 & 0.796 \\ \hdashline %
  LR & 0.781 & 0.754 & 0.837 \\ %
\end{tabular}
 \caption{Accuracy of the different models on our dataset using cross validation with k=$5$. {SciBERT}$^f$ outperforms.}
  \label{tab:our_dataset}
\end{table}

\begin{table}
 \begin{tabular}{l|c|c|c}
  Model & Accuracy & Precision & Recall \\ 
  \hline 
  {\MLP}  & 0.884 & 0.913 & 0.848 \\ \hdashline %
  SciBERT & 0.882 & 0.909 & 0.848 \\ \hdashline %
  \makecell[l]{{SciBERT}$^f$} & \textbf{0.903} & \textbf{0.921} & \textbf{0.882} \\ \hdashline %
  BERT & 0.863 & 0.905 & 0.810 \\ \hdashline %
  \makecell[l]{{BERT}$^f$} & \textbf{0.903} & \textbf{0.921} & \textbf{0.882} \\ \hdashline %
  RF & 0.713 & 0.708 & 0.725 \\ \hdashline %
  LR & 0.765 & 0.755 & 0.787 \\ %
\end{tabular}
 \caption{Accuracy of the different models on our Ig-Nobel retrieval test set. The combination of SOTA pretrained models and our features is superior.}
  \label{tab:ig_ret}
\end{table}

\section{Evaluation ``in the Wild''}
\label{sec:SSExp}
\begin{table*}[h!]
\begin{tabular}{p{11.1cm}|c}
Title & Models \\
\hline
    \makecell[l]{The kinematics of eating with a spoon: Bringing the food to the mouth,\\or the mouth to the food?} & \makecell{\MLP, {BERT}$^f$, {SciBERT}$^f$}\\  \hdashline %
    Do bonobos say NO by shaking their head? & \makecell{\MLP, {BERT}$^f$, {SciBERT}$^f$}\\  \hdashline %
    Is Anakin Skywalker suffering from borderline personality disorder? & \makecell{\MLP, {BERT}$^f$, {SciBERT}$^f$}\\  \hdashline %
    Not eating like a pig: European wild boar wash their food & \makecell{\MLP, {BERT}$^f$}\\  \hdashline %
    Why don't chimpanzees in Gabon crack nuts? & \makecell{{SciBERT}$^f$, {BERT}$^f$}\\ \hdashline %
    {Why do people lie online? ``Because everyone lies on the internet"} & {BERT}$^f$\\  \hdashline
    Which type of alcohol is easier on the gut? & {BERT}$^f$\\  \hdashline
    Rainbow connection and forbidden subgraphs & {BERT}\\  \hdashline
    A scandal of invisibility: making everyone count by counting everyone & {SciBERT}\\  \hdashline
    Where do we look when we walk on stairs? Gaze behaviour on stairs, transitions, and handrails & {SciBERT}
\end{tabular}
\caption{A sample of top rated papers found by our models.}
 \label{tab:sample}
\end{table*}

Our main motivation in this work is to recommend papers worthy of an Ig Nobel prize. In this section, we test our models in a more realistic setting; we run them on a large sample of scientific papers, ranking each paper according to their certainty in the label (`humorous'), and identifying promising candidates. 
We use the same dataset of $630$k papers from Semantic Scholar used for training the LMs (Section \ref{sec:analysis}). We compute funniness according to our models (excluding random forest and logistic regression, which performed poorly). Table \ref{tab:sample} shows examples of top-rated titles. We use the Amazon Mechanical Turk (MTurk) crowdsourcing platform to assess models' performance.

In an exploratory study, we asked people to rate the funniness of titles on a Likert scale of $1$-$5$. We noted that people tended to confuse funny \emph{research topic} and funny \emph{title}. For example, titles like ``Are you certain about SIRT?'' or ``NASH may be trash'' received high funniness scores, even though the research topic is not even clear from the title.
To mitigate this problem, we redesigned the study to include \emph{two} $5$-point Likert scale questions: 1) whether the \textit{title} is funny, and 2) whether the \textit{research topic} is funny. 
This addition seems to indeed help workers understand the task better. Example papers rated as serious title, funny topic include ``Hat-wearing patterns in spectators attending baseball games: a $10$-year retrospective comparison''. Funny title, serious topic include ``Slicing the psychoanalytic pie: or, shall we bake a new one? Commentary on Greenberg''. Unless stated otherwise, the evaluation in the reminder of the paper was done on the ``funny topic'' Likert scale.

We paid crowd workers $\$0.04$ per title. As this task is challenging, we created a qualification test with $4$ titles ($8$ questions), allowing for one mistake. The code for task and test can be found in the repository\footref{github}. We also required workers to have completed at least $1$,$000$ approved HITs with at least $97\%$ success rate. 

All algorithms classified and ranked (according to certainty) all $630$k papers. However, in any reasonable use-case, only the top of the ranked list will ever be examined. There is a large body of work, both in academia and industry, studying how people interact with ranked lists (in particular, search result pages) \citep{kelly2015many, Johannes-Google-CTR}.
Many information retrieval algorithms assume the likelihood of the user examining a result to exponentially decrease with rank. 
The conventional wisdom is that users rarely venture into the second page of search results. 

Thus, we posit that in our scenario of Ig Nobel recommendations, users will be willing to read only the several tens of results. We choose to evaluate the top-$300$ titles for each of our five models, to study (in addition to the performance at the top of the list) how performance decays. 
We also included a baseline of $300$ randomly sampled titles from Semantic Scholar.
Altogether we evaluated $1375$ titles (due to overlap). Each title was rated by five crowd workers. 
Overall, $13$ %
different workers passed our test. Seven workers annotated less than $300$ titles, while four annotated above $1$,$300$ each. 



\xhdr{Decision rule}
Each title was rated by five different crowd workers on a $1$-$5$ scale. There are several reasonable ways to aggregate these five continuous scores to a binary decision. A commonly-used aggregation method is the majority vote. 
The majority vote should return the clear-cut humorous titles. However, we stress that humor is very subjective (and in the case of scientific humor, quite subtle). Indeed, annotators had low agreement on the topic question (average pairwise Spearman $\rho = 0.27$). 

Thus, we explored more aggregation methods\footnote{For completeness, see Figure \ref{fig:Pre@kMajority} in Appendix \ref{app:resultsMajority}.}. Our hypothesis class is of the general form ``at least $k$ annotators gave a score at least $m$'' \footnote{There is a long-running debate about whether it is valid to average Likert scores. We believe we cannot treat the ratings in this study as interval data.}. 
To pick the best rule, we conducted two exploratory experiments: 
In the first one, we recruited an expert scientist and thoroughly trained him on the problem. He then rated $90$ titles and we measured the correlation of different aggregations with his ratings. Results are summarized in table \ref{tab:MTurkVal}: The highest-correlation aggregation is when at least one annotator crossed the $3$ threshold (Spearman $\rho = 0.7$). 

\begin{table}[t!]
    \begin{tabular}{l|c|c|c}
        \makecell{Decision\\rule} & Threshold & \makecell{Expert\\corr.} & \makecell{Labeled data\\accuracy} \\
        \hline
        \multirow{2}{*}{\makecell{Min. $1$\\annotator}} & $3$ & \textbf{0.7} & \textbf{0.84} \\
        \cdashline{2-4} & $4$ & 0.49 & 0.83\\
        \hline
        \multirow{2}{*}{\makecell{Min. $2$\\annotators}} & $3$ & 0.47 & 0.82\\
        \cdashline{2-4} & $4$ & 0.19 & 0.73\\
        \hline
        \multirow{2}{*}{\makecell{Min. $3$\\annotators}} & $3$ & 0.15 & 0.78\\
        \cdashline{2-4} & $4$ & 0.02 & 0.62\\
    \end{tabular}
    \caption{Spearman correlation of MTurk annotators with our expert, along with accuracy of MTurk annotators on our labeled dataset for the various mapping methods of the form ``minimum (min.) $k$ annotators gave a score at least $m$ (threshold)''.}
    \label{tab:MTurkVal}
\end{table}

In the second experiment, we used the exact same experimental setup as the original task, but with \emph{labeled} data. We used $100$ Ig Nobel winners as positives and a random sample of $100$ papers as negatives. The idea was to see how crowd workers rate papers that we know are funny (or not). Table \ref{tab:MTurkVal} shows the accuracy of each aggregation method. Interestingly, the highest accuracy is achieved with the same rule as in the first experiment (at least one crossing $3$). Thus, we chose this aggregation rule. 

We believe the method outlined in this section could be more broadly applicable to aggregation of crowd sourced annotations for subjective questions.

\begin{figure*}[t!]
\centering
     \includegraphics[width=0.65\linewidth]{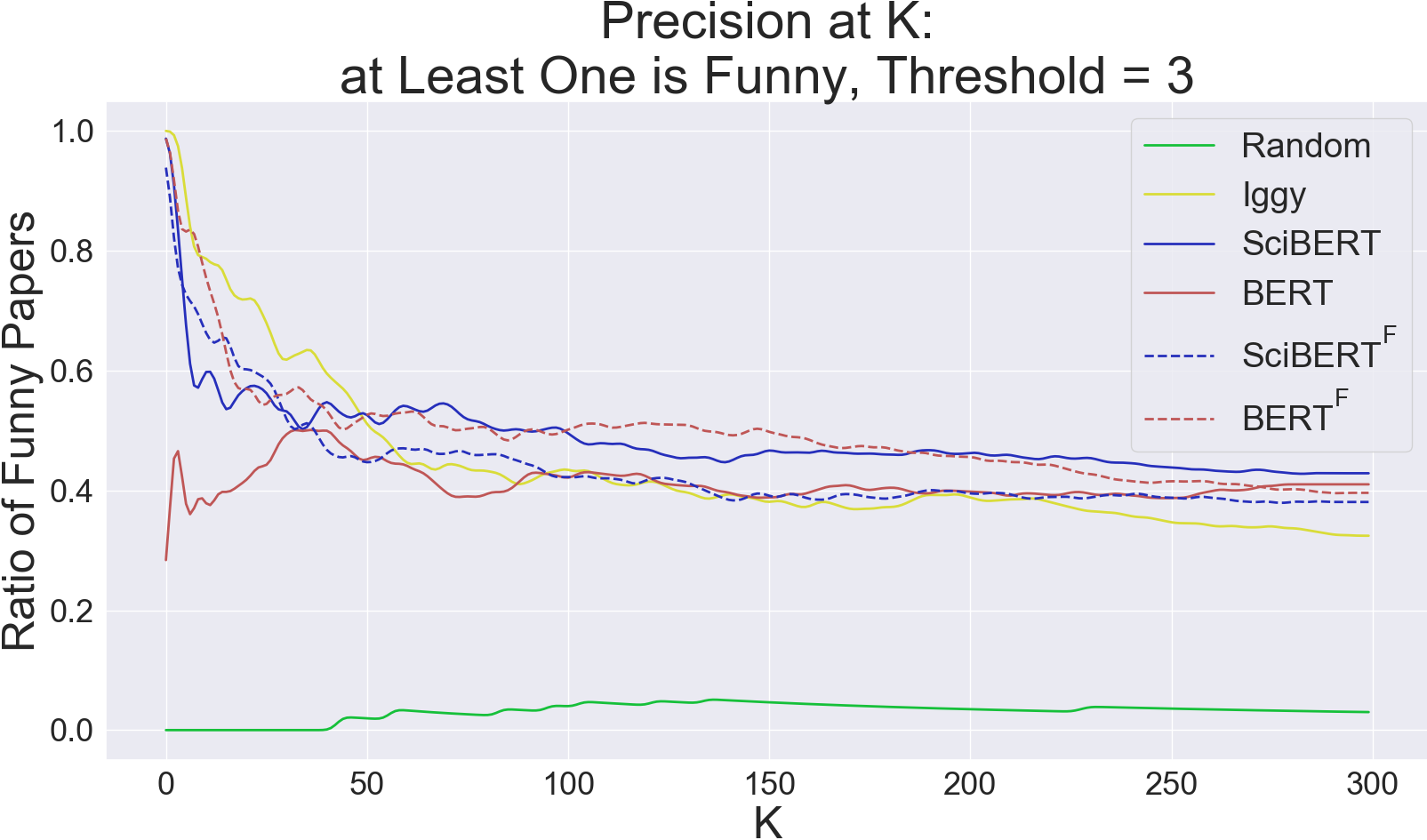}
     \caption{Precision at k for our chosen decision rule. \MLP outperforms the other models for $0 < k < 50$. For larger $k$, {SciBERT}$^f$ and BERT achieve better precision.}
     \label{fig:pre@k1}
\end{figure*}

\xhdr{Results}
%
Figure \ref{fig:pre@k1} shows precision at $k$ for the top-rated $300$ titles according to each model. 
The random baseline is $\sim 0.03$. Upon closer inspection, these seem to be false positives of the annotation.

We have argued that in our setting it is reasonable for users to read the first several tens of results. In this range, \MLP slightly outperforms the other four models (BERT is particularly bad, as it picks up on short, non-informative titles). For larger $k$ values {SciBERT} and BERT$^f$ take the lead. 
We note that even at $k=300$, all models still achieve considerable (absolute) precision.


We obtain similar results using normalized discounted cumulative gain (nDCG), a common measure for ranking quality (see Table \ref{tab:pre@k} for nDCG scores for the top $50$ and the $300$ papers). Overall, these relatively high scores suggest that our models are able to identify funny papers. 

\begin{table}
 \begin{tabular}{l|m{6em}|m{6em}}
  Model & \makecell{Precision\\at k=$50$} & \makecell{Precision\\at k=3$00$}\\ 
  \hline 
  {\MLP}  &\ \ \ \ \ \ \ \  \textbf{ 0.6} &\ \ \ \ \ \ \ \  0.37 \\ \hdashline %
  SciBERT &\ \ \ \ \ \ \ \  0.57 &\ \ \ \ \ \ \ \   \textbf{0.46} \\ \hdashline %
  \makecell[l]{{SciBERT}$^f$} &\ \ \ \ \ \ \ \   0.53 &\ \ \ \ \ \ \ \   0.41 \\ \hdashline %
  BERT &\ \ \ \ \ \ \ \    0.44 &\ \ \ \ \ \ \ \   0.41 \\ \hdashline %
  \makecell[l]{{BERT}$^f$} &\ \ \ \ \ \ \ \   {0.58} &\ \ \ \ \ \ \ \   {0.43} \\ \hdashline %
\end{tabular}
 \caption{Precision at k of our models on the Semantic Scholar corpus for k=$\{50, 300\}$. These relatively high scores suggest that our models are able to identify funny papers.}
  \label{tab:pre@k}
\end{table}

We stress that \MLP is a small and simple network ($\sim 33$k parameters), compared to pretrained $110$ \textit{million} parameters BERT-based models. 
Yet despite its simplicity, {\MLP}'s performance is roughly comparable to BERT-based methods. We believe this demonstrates the power of implementing insights from domain experts. 
We hypothesize that if the fine-tuning dataset were larger, {BERT}$^f$ and {SciBERT}$^f$ would outperform the other models.

\section{Analysis}
\label{sec:interp}
\subsection{Importance of Literature-based Features}
\label{sec:featureImportance}

Taking a closer look at the actual papers in the experiment of Section \ref{sec:SSExp}, the overlap between the three \textit{feature-based} models is $26-56\%$ (for $1<k<50$) and $39-62\%$ (for $1<k<300$). BERT had very low overlaps with all other models ($0\%$ in top 50, $10\%$ in all 300). SciBERT had almost no overlap in top 50 (maximum $2\%$), $10-40\%$ in all 300 (see full details in Appendix \ref{app:overlap}). We believe this implies that the features were indeed important and informative for both {BERT}$^f$ and {SciBERT}$^f$.

\subsection{Interpreting \MLP}
\label{sec:SHAP}

We have seen {\MLP} performs surprisingly well, given its relative simplicity. In this section, we wish to better understand the reasons. We chose to analyze \MLP with Shapely additive explanations (SHAP) \citep{lundberg2017unified}. SHAP is a feature attribution method to explain the output of any black-box model, shown to be superior to more traditional feature importance methods. Importantly, SHAP provides insights both globally and locally (i.e., for specific data points). 


\xhdr{Global interpretability} We compute feature importance globally. 
Among top contributing features we see multiple features corresponding to incongruity (both alone and combined with funniness) and to word/sentence simplicity. Interestingly, features based on the one-liner jokes seem to play an important role (See Figure \ref{fig:SHAPVarImportance} in Appendix \ref{app:shapFigures}).

\xhdr{Local interpretability} 
To understand how \MLP errs, we examined the SHAP decision plots for false positives and false negatives (See Figure \ref{fig:SHAPDecisionFNFP} in Appendix \ref{app:shapFigures}). These show the contribution of each feature to the final prediction for a given title, and thus can help ``debugging'' the model. 

Looking at false negatives, it appears that various perplexity features misled \MLP, while funniness and joke LM steered it in the right direction.  
We see a contrary trend in false positives: perplexity helped, and joke LM confused the classifier. 

We also observe that the model learned that a long title is an indication of a serious paper. We expected our rudeness classifier to play a bigger role in some of the titles (e.g., ``Adaptive inter-population differences in blue tit life-history traits on Corsica''), but the signal was inconclusive, perhaps indicating our rudeness classifier is lacking.






\subsection{Observations}



We now take a more qualitative approach to understand the models.
First, we set out to explore whether the models confuse funny titles and funny topics. 
Using the crowd sourced annotations from Section \ref{sec:SSExp}, we measure the portion of this mistake in the top-rated $300$ titles of all five models. That is, we check in how many cases our models classify a title as ``Ig Nobel-worthy'' while the workers have classified it as ``funny title and non-funny topic''.
\MLP had the highest degree of such confusion ($0.28$). Similarly, {BERT}$^f$ and {SciBERT}$^f$ exhibit more confusion than the versions without features ($0.24$, $0.19$ compared to $0.13$, $0.08$). Random baseline is $0.02$.
Examples of this kind of error include ``A victim of the Occam's razor.'', ``While waiting to buy a Ferrari, do not leave your current car in the garage!'', and ``Reinforcement learning: The good, the bad and the ugly?''. All were classified as Ig Nobel-worthy, although their topic is serious (or even unclear from the title).  

Looking closer at the data, we observe that a high portion of these are \emph{editorials} with catchy titles. As our dataset does not differentiate between editorials and real research contributions, filtering editorials is not straightforward. Interestingly, the portion of editorials is also greater in the lowest annotators' agreement area, hinting that this confusion also occurs in humans.

In addition to editorials, we notice another category of papers causing the same type of confusion. There are papers dealing with disturbing or unfortunate topics (violence, death, sexual abuse), whose titles include literary devices used to lighten the mood. Censored (for the readers' own well-being) examples include ``Licorice for hepatitis C: yum-yum or just ho-hum?'', ``The song of the siren: Dealing with masochistic thoughts and behaviors''. 


\xhdr{A note on scientific disciplines}
Another observation we make concerns with the portion of Ig Nobel-worthiness across the different scientific disciplines. We notice that most papers classified by our models as funny belong to social sciences (``Dogs can discriminate human smiling faces from blank expressions'') or medicine (``What, if anything, can monkeys tell us about human amnesia when they can't say anything at all?''), compared to exact sciences (``The kinematics of eating with a spoon: bringing the food to the mouth, or the mouth to the food?''). 
We believe this might be the case since, quite often, social sciences and medicine papers study topics that are more familiar to the layperson. 
We also note that although our models performed about the same across the different disciplines, they were slightly better in psychology. 





\section{Conclusions \& Future Work}
\label{sec:conc}


In this work, we presented a novel task in humor recognition -- detecting funny and unusual scientific papers, which represents a subtle and sophisticated humor type. It has important characteristics (short, simple syntax, stand-alone) making it a (relatively) clean setting to explore computational humor.




We created a dataset of funny papers and constructed models, distilling humor literature into features as well as harnessing SOTA advances in NLP. 
We conducted experiments both on our dataset and in a real-world setting, identifying funny papers in a corpus of over $0.6$M papers. All models were able to identify funny papers, achieving high nDCG scores. Interestingly, despite the simplicity of the literature-based \MLP, its performance was overall comparable to complex, BERT-based models.

Our dataset can be further used for various humor related tasks. For example, it is possible to use it to create an \emph{aligned} corpus, pairing every funny paper title with a nearly identical but serious  title, using methods similar to \citet{west2019reverse}. 
 This would allow us to understand why a paper is funny at a finer granularity, by identifying the exact words that make the difference. This technique will also allow exploring different types of ``funny''. 
 
 Another possible use of our dataset is to collect additional meta-data about the papers (e.g., citations, author information) to explore questions about whether funny science achieves disproportionate attention and engagement, who tends to produce it (and at which career stage), with implications to science of science and science communication.
 
Another interesting direction is to expand beyond paper titles and consider the paper abstract, or even full text. This could be useful in examples such as the Ig Nobel winner ``Cure for a Headache'', which takes inspiration from woodpeckers to help cure headaches in humans. 
%

Finally, we believe multi-task learning is a direction worth pursuing towards creating a more {holistic and robust humor classifier}. In multi-task learning, the learner is challenged to solve multiple problems at the same time,  often resulting in better generalization and better performance on each individual task \citep{Ruder2017AnOO}. As multi-task learning enables unraveling cross-task similarities, we believe it might be particularly fruitful to apply to tasks highlighting different aspects of humor. We believe our dataset, combined with other task specific humor datasets, could assist in pursuing such a direction.




Despite the tongue-in-cheek nature of our task, we believe that 
computational humor has tremendous potential to create personable interactions, and can greatly contribute to a range of NLP applications, from chatbots to educational tutors. We also wish to promote complementing data-driven research with insights from more-traditional fields. We believe combining such insights could, in addition to improving performance, enrich our understanding of core aspects of being human.


\section*{Acknowledgments} 

We thank the reviewers for their insightful comments. 
We thank Omri Abend, Michael Doron and Meirav Segal, Ronen Tamari and Moran Mizrahi for their help, and Shuki Cohen for preliminary discussions. This work was supported by the European Research Council (ERC) under the European Union's Horizon 2020 research and innovation programme (grant no. 852686, SIAM), US National
Science Foundation, US-Israel Binational Science Foundation (NSF-BSF) grant no. 2017741, and Amazon Research Awards.

\bibliographystyle{acl_natbib}
\bibliography{acl2021}

\appendix
\clearpage 

\begin{figure*}[t!]
\centering
    \begin{subfigure}[b]{0.43\textwidth}
         \includegraphics[width=\textwidth]{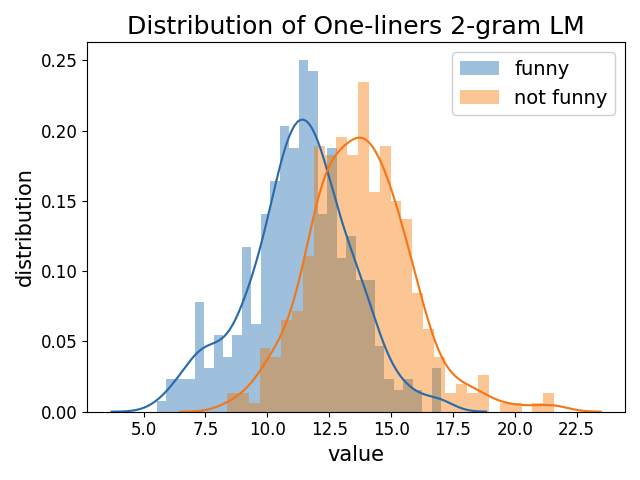}
     \end{subfigure}
    \begin{subfigure}[b]{0.43\textwidth}
         \includegraphics[width=\textwidth]{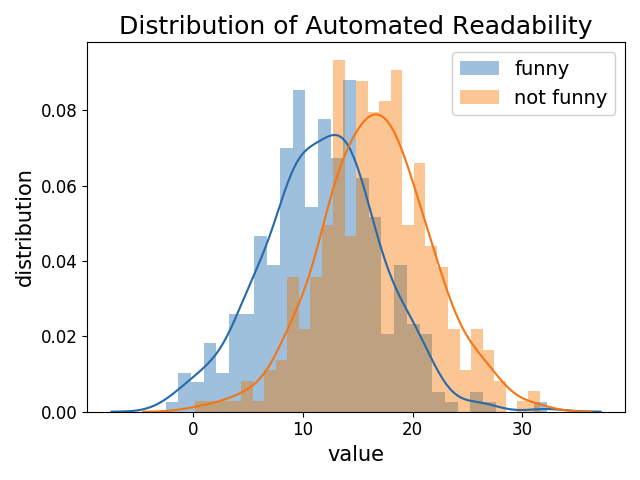}
     \end{subfigure}
  \caption{Distribution and Gaussian fit of two representative features: one-liners 2-gram LM mean perplexity (left) and automated readability index (right), indicating the predictive power of these features.}
\label{fig:DistFeatures}
\end{figure*}

\section{Supplementary figures}

\subsection{Dataset Analysis}
\label{app:dataAnalysis}

In Section \ref{sec:analysis} we presented $127$ humor literature-based features. Here we present the distribution of two example features in funny vs. serious papers in our dataset (described in Section \ref{sec:dataset}). These examples represent the general trend, as many features show predictive power (see Figure \ref{fig:DistFeatures}).

\subsection{``In the Wild'' Study Results}
\label{app:resultsMajority}

\begin{figure*}
\centering
    \includegraphics[width=0.6\linewidth]{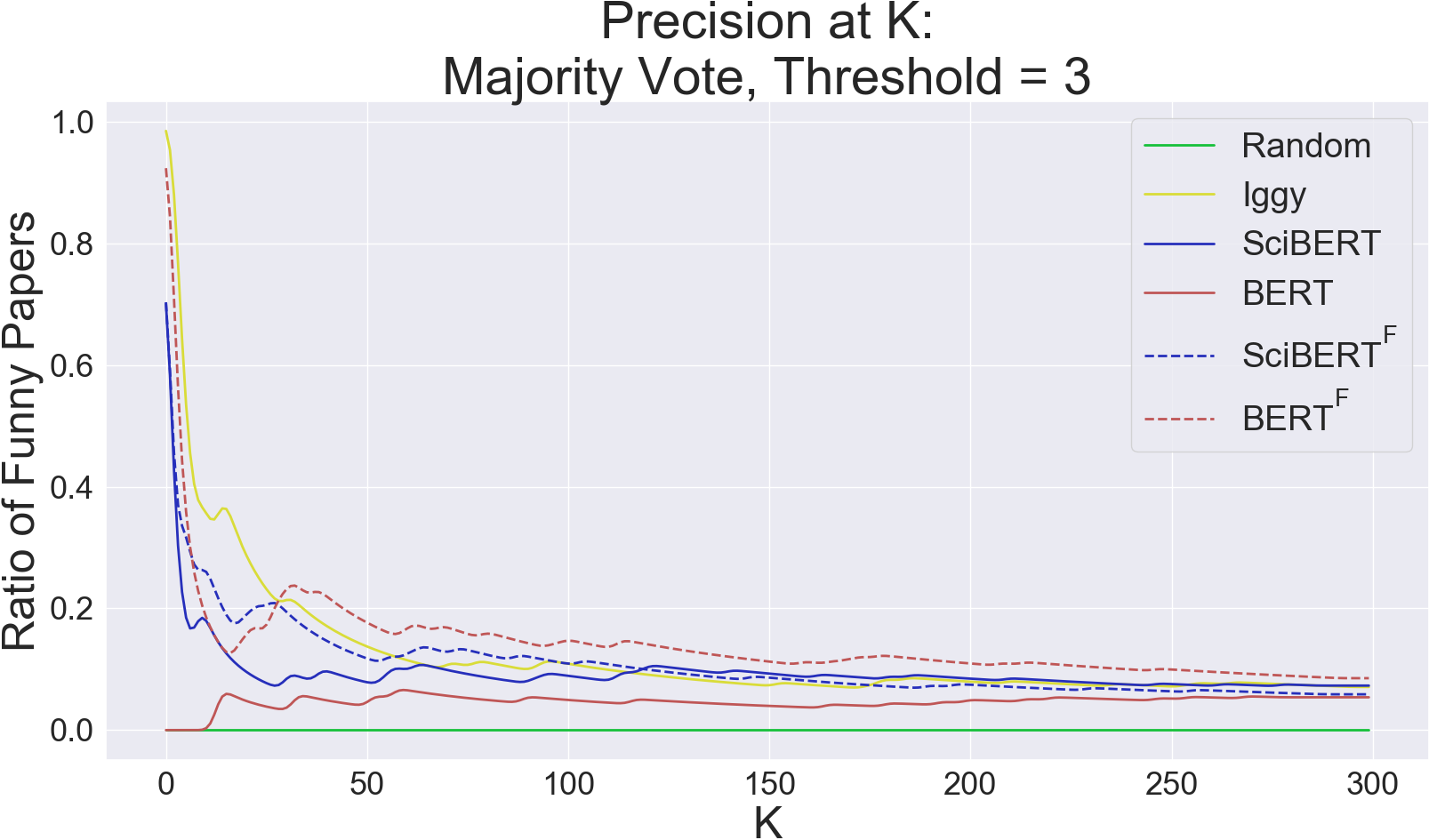}
     \caption{Precision at k for majority vote. {\MLP} outperforms until $k=30$ and $\textrm{BERT}^f$ takes the lead afterwards.}
     \label{fig:Pre@kMajority}
 \end{figure*}
 
For the ``in the wild'' evaluation executed using Semantic Scholar data, we used crowdsourcing annotations (see Section \ref{sec:SSExp}).
Each title was rated by five different crowd workers on a $1$-$5$ scale, while our models provide binary decision. There are several reasonable ways to aggregate these five continuous scores to a binary decision. We  choose a rule in a data-driven manner (see ``Decision rule'' in Section \ref{sec:SSExp}). For completeness, here we show the commonly-used aggregation method of \emph{majority}. We show here the precision at $k$ of our five models using the majority vote aggregation rule with a cutoff at $3$ (see Figure \ref{fig:Pre@kMajority}). {\MLP} outperforms until $k=30$, where $\textrm{SciBERT}^f$ takes the lead afterwards. 
\subsection{Models' Overlap}
\label{app:overlap}

\begin{table*}
\centering
\begin{tabular}{l|c|c|c|c}
& \MLP & {BERT}$^f$ & {SciBERT}$^f$ & BERT \\
\hline 
$\textrm{BERT}^f$  & $0.56\ \| \ 0.62$ & & & \\ \hline%
$\textrm{SciBERT}^f$ & $0.26\ \| \ 0.39$ & $0.36\ \| \ 0.48$ & & \\  \hline%
BERT & $0\ \| \ 0$ & $0\ \| \ 0.1$ & $0\ \| \ 0.1$ & \\  \hline%
SciBERT & $0\ \| \ 0.3$ & $0.02\ \| \ 0.4$ & $0.02\ \| \ 0.2$ & $0\ \| \ 0.1$
\end{tabular}
\caption{Models' overlap for the top rated $50$ and $300$ (left number in a cell corresponds to the overlap in the top $50$ and right number corresponds to the $300$). The overlap between the $3$ \textit{features-based} models was found to be high compared with BERT and SciBERT. We believe this implies that the features were indeed important for our SOTA based models, {BERT}$^f$ and {SciBERT}$^f$.}
\label{tab:overlap}
\end{table*}

In Section \ref{sec:featureImportance} we discuss the importance of our literature-based features by showing that models who received them as input indeed found them useful. The overlap was measured on the top $50$ and top $300$ papers retrieved using our five models on the Semantic Scholar data (see Section \ref{sec:SSExp} for the full experimental setup).
The overlap between the $3$ \textit{features-based} models was found to be high (see Table \ref{tab:overlap}). Both BERT and SciBERT had very low overlaps with all other models. We believe this implies that the features were indeed important for our SOTA based models, {BERT}$^f$ and {SciBERT}$^f$.

\subsection{SHAP Analysis}
\label{app:shapFigures}

\begin{figure*}[h!]
  \centering
  \includegraphics[width=\linewidth]{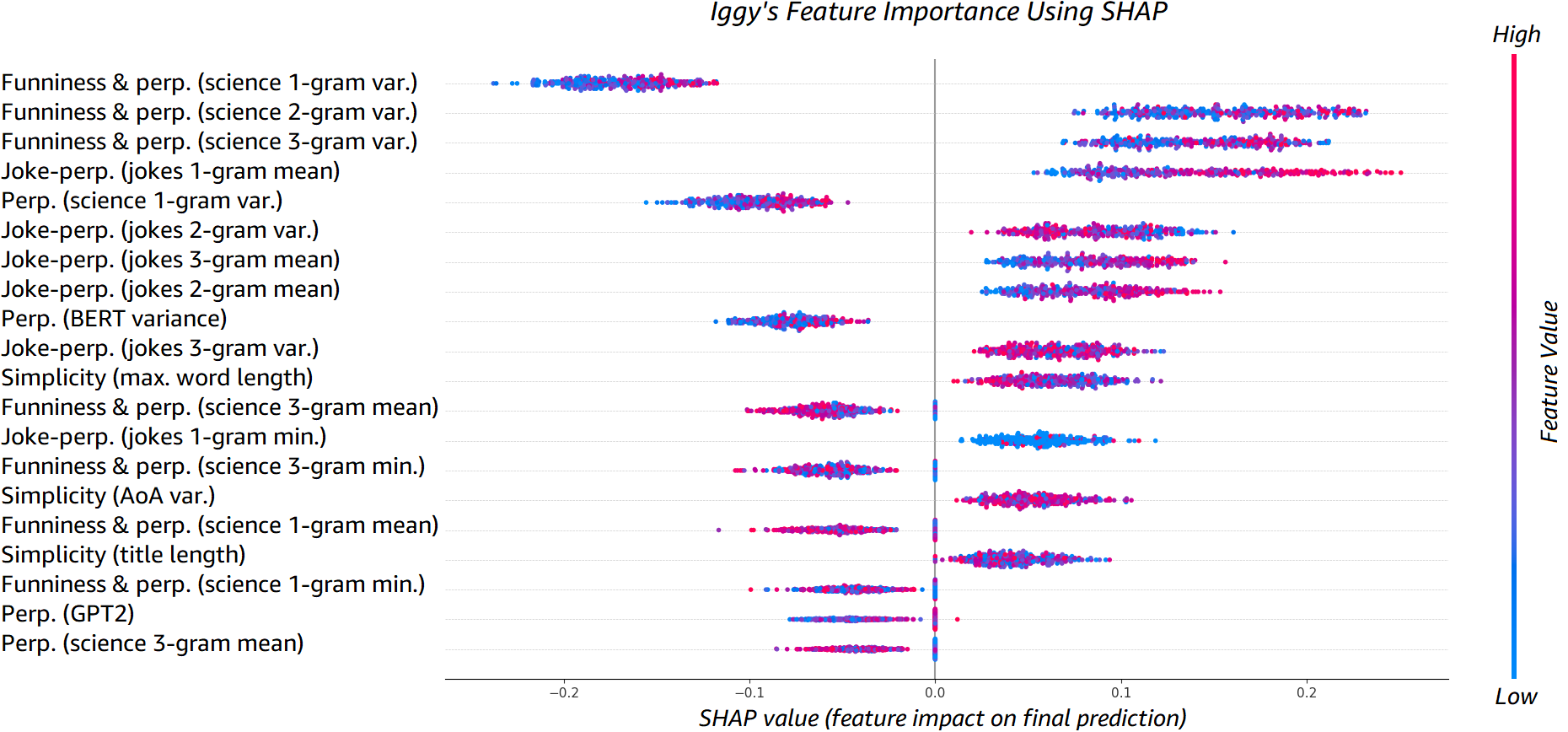}
  \caption{Feature importance SHAP analysis done on the {\MLP} model. The top variables according to this plot contribute more than the bottom ones (have high predictive power). The analysis reveals that the highest contribution corresponds to short, funny, and simple words (where simplicity was measured using features such as AoA and readability). We also notice that features which are based on the one-liners LMs contributed much to the final prediction, meaning that there is indeed some similarity between funny titles and short jokes.}
\label{fig:SHAPVarImportance}
\end{figure*}

\begin{figure*}
     \centering
     \begin{subfigure}[b]{\textwidth}
         \centering
         \includegraphics[width=\textwidth]{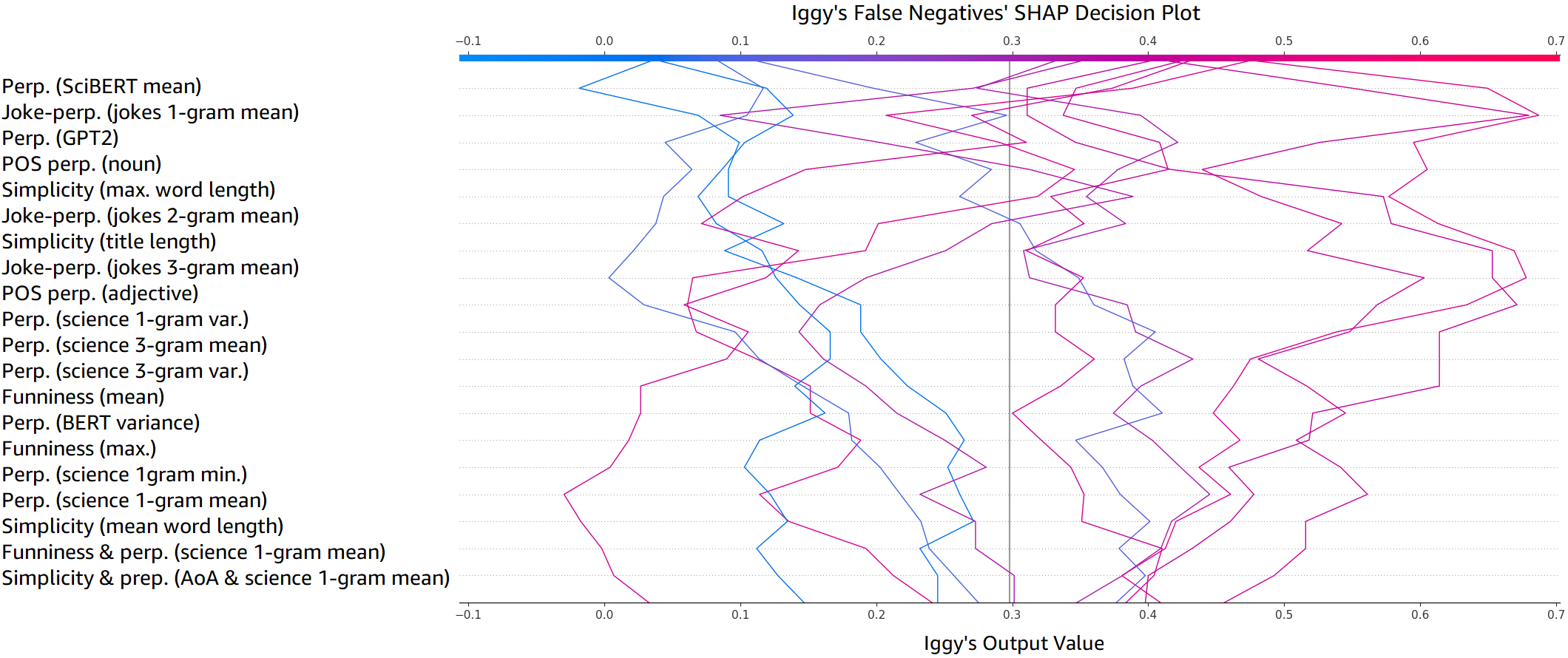}
         \caption{SHAP decision plot for the $12$ false negative of Iggy from our test set. Perplexity features misled \MLP, while funniness and joke LM ones provided informative input.}
     \end{subfigure}
     \hfill
     \begin{subfigure}[b]{\textwidth}
         \centering
         \includegraphics[width=\textwidth]{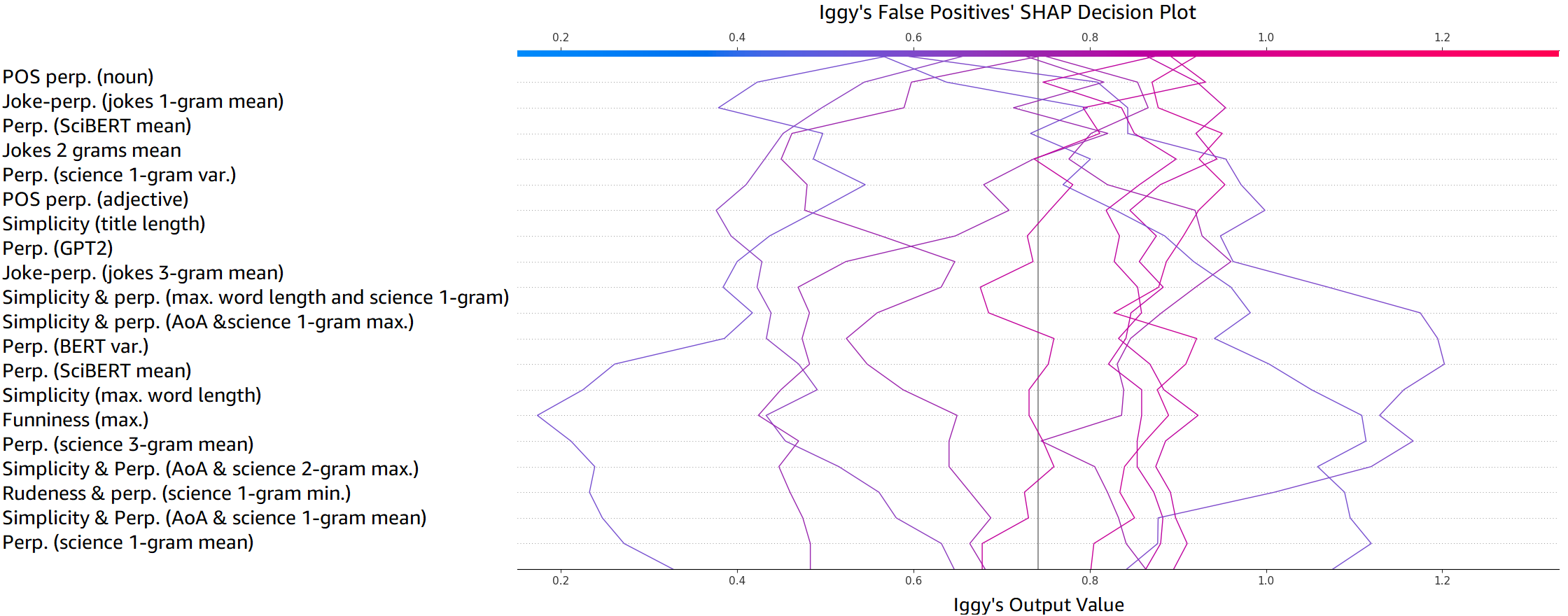}
         \caption{SHAP decision plot for the $11$ false positive of Iggy from our test set. Perplexity helped shifting the output towards the correct label, joke LM features confused the classifier}
     \end{subfigure}
     \caption{SHAP decision plot for {\MLP}'s false negatives and positives from our test set. Decision plots show the contribution of each feature to the final prediction for a given data point. Starting at the bottom of the plot, the prediction line shows how the SHAP values (i.e., the feature effects) accumulate to arrive at the model’s final score at the top of the plot. To get a better intuition, one can think of it in terms of a linear model where the sum of effects, plus an intercept, equals the prediction.}
      \label{fig:SHAPDecisionFNFP}    
\end{figure*}

In Section \ref{sec:SHAP} we analysed \MLP using SHAP \citep{lundberg2017unified}. 
We compute feature importance globally (Figure \ref{fig:SHAPVarImportance}). 
To understand how \MLP errs, we examined the SHAP decision plots for false positives and false negatives (Figure \ref{fig:SHAPDecisionFNFP}). 
Decision plots show the contribution of each feature to the final prediction for a given title. Thus, it can help ``debugging'' the model's mistakes.

\section{Reproducibility}

\subsection{Code and Data Availability}

Dataset, code, and data files can be found in our Github repository\footref{github}. 

\section{Implementation details}
\label{app:implementation_details}

\subsection{Fine-Tuning GPT-2 LM}
\label{app:GPT2}
To fine-tune GPT-2 we used Huggingface's Transformers package\footnote{\rurl{huggingface.co/transformers/}}. We fine-tuned the model using learning rate $=5$e$-5$, one epoch, batch size of 4, weight decay $=0$, max gradient norm $=1$ and random seed $=42$. Optimization was done using Adam with epsilon $=1$e$-8$. Model configurations were set to default.

\subsection{{\MLP}  Classifier}
\label{app:MLP}
We used a simple MLP  with a single hidden layer of $256$ neurons. We trained the MLP until convergence, using Adam optimizer, a learning rate of $0.001$ and an L$2$ penalty of $2$.

\subsection{Fine-tuning SciBERT \& BERT}
\label{app:finetune_bert}

To fine-tune SciBERT \& BERT we used Huggingface's Transformers package. We fine-tuned both models with learning rate $=5$e$-5$ for $3$ epochs with batch size of $32$, maximal sequence length of $128$ and random seed $=42$. Optimization was done using Adam with warm-up $=0.1$ and weight decay of $0.01$ Model configurations were set to default.

\subsection{{SciBERT}$^f$ \& {BERT}$^f$ Models}
\label{app:combined_model}

\begin{figure*}[t!]
  \centering
  \includegraphics[width=0.6\linewidth]{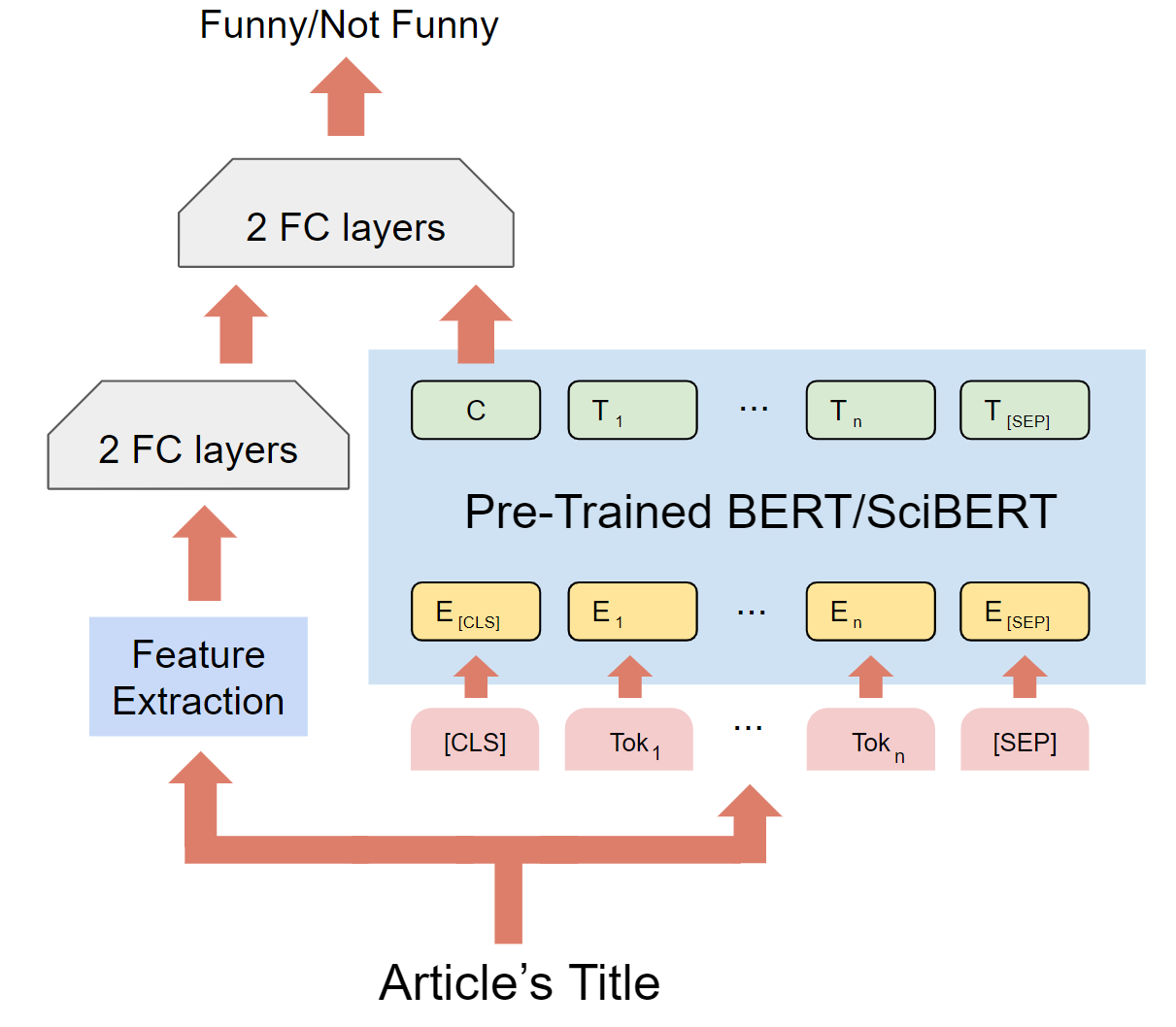}
  \caption{The flow of {SciBERT}$^f$ \slash {BERT}$^f$. A 2-layers MLP recieves an input the concatenation of two vectors: our features' embedding and the last hidden vector ([CLS]) from BERT \slash SciBERT.}
\label{fig:modelflow}
\end{figure*}

As specified in Section \ref{sec:model}, these models were constructed as follows (see Figure \ref{fig:modelflow}). Each model had two inputs -- the raw text of the title, and a vector of our $127$ features. The feature vector is fed to an MLP with a single hidden layer of $512$ neurons and an output size of $512$ neurons as well. The raw text is fed to a \textit{frozen} SciBERT \slash BERT model. We collect the last hidden vector ([CLS]) from BERT \slash SciBERT. Next, we concatenate this vector to the output of the features-MLP network and pass the result to a second MLP with a single hidden layer of $1$,$024$ neurons. The output of this MLP, then, is fed to a Softmax layer, which represents the final prediction of the model. 

We train the model using a cross-entropy loss and the same parameters that were used to train the vanilla SciBERT \slash BERT model. Those parameters are described in Appendix \ref{app:finetune_bert}.



\end{document}